\documentclass{article}
\usepackage{ijcai20}

\usepackage{times}
\usepackage{soul}
\usepackage{url}
\usepackage[hidelinks]{hyperref}
\usepackage[utf8]{inputenc}
\usepackage[small]{caption}
\usepackage{graphicx}
\usepackage{amsmath}
\usepackage{amsthm}
\usepackage{booktabs}
\usepackage{algorithm}
\usepackage{algorithmic}
\usepackage{multirow}
\usepackage{subfigure}
\usepackage[table,xcdraw]{xcolor}
\usepackage[final]{changes}
  \definechangesauthor[name={feng}, color=orange]{feng}

\title{``Wait, I'm Still Talking!'' \\ Predicting the Dialogue Interaction Behavior\\Using Imagine-Then-Arbitrate Model}

\author{
Zehao Lin$^1$
\and
Xiaoming Kang$^2$\and
Guodun Li$^1$\And
Feng Ji$^2$\and
Haiqing Chen$^2$\and
Yin Zhang$^1$
\affiliations
$^1$College of Computer Science and Technology, Zhejiang University\\
$^2$DAMO Academy, Alibaba Group\\
\emails
georgelin@zju.edu.cn,
kxm180043@alibaba-inc.com,
guodun.li@gmail.com,
\{zhongxiu.jf, haiqing.chenhq\}@alibaba-inc.com,
zhangyin98@zju.edu.cn
}

\begin{document}

\maketitle

\begin{abstract}
Producing natural and accurate responses like human beings is the ultimate goal of intelligent dialogue agents. So far, most of the past works concentrate on selecting or generating one pertinent and fluent response according to current query and its context. These models work on a one-to-one environment, making one response to one utterance each round. However, in real human-human conversations, human often sequentially sends several short messages for readability instead of a long message in one turn. Thus messages will not end with an explicit ending signal, which is \deleted{unnecessary for human but} crucial for agents to decide when to reply. So the first step for an intelligent dialogue agent is not replying but deciding if it should reply at the moment. To address this issue, in this paper, we propose a novel Imagine-then-Arbitrate (ITA) neural dialogue model to help the agent decide whether to wait or to make a response directly. Our method has two imaginator modules and an arbitrator module. The two imaginators will learn the agent's and user's speaking style respectively, generate possible utterances as the input of the arbitrator, combining with dialogue history. And the arbitrator decides whether to wait or to make a response to the user directly. To verify the performance and effectiveness of our method, we prepared two dialogue datasets and compared our approach with several popular models. Experimental results show that our model performs well on addressing ending prediction issue and outperforms baseline models.
\end{abstract}

\section{Introduction}
\begin{figure}
\centering
\includegraphics[width=0.7\linewidth]{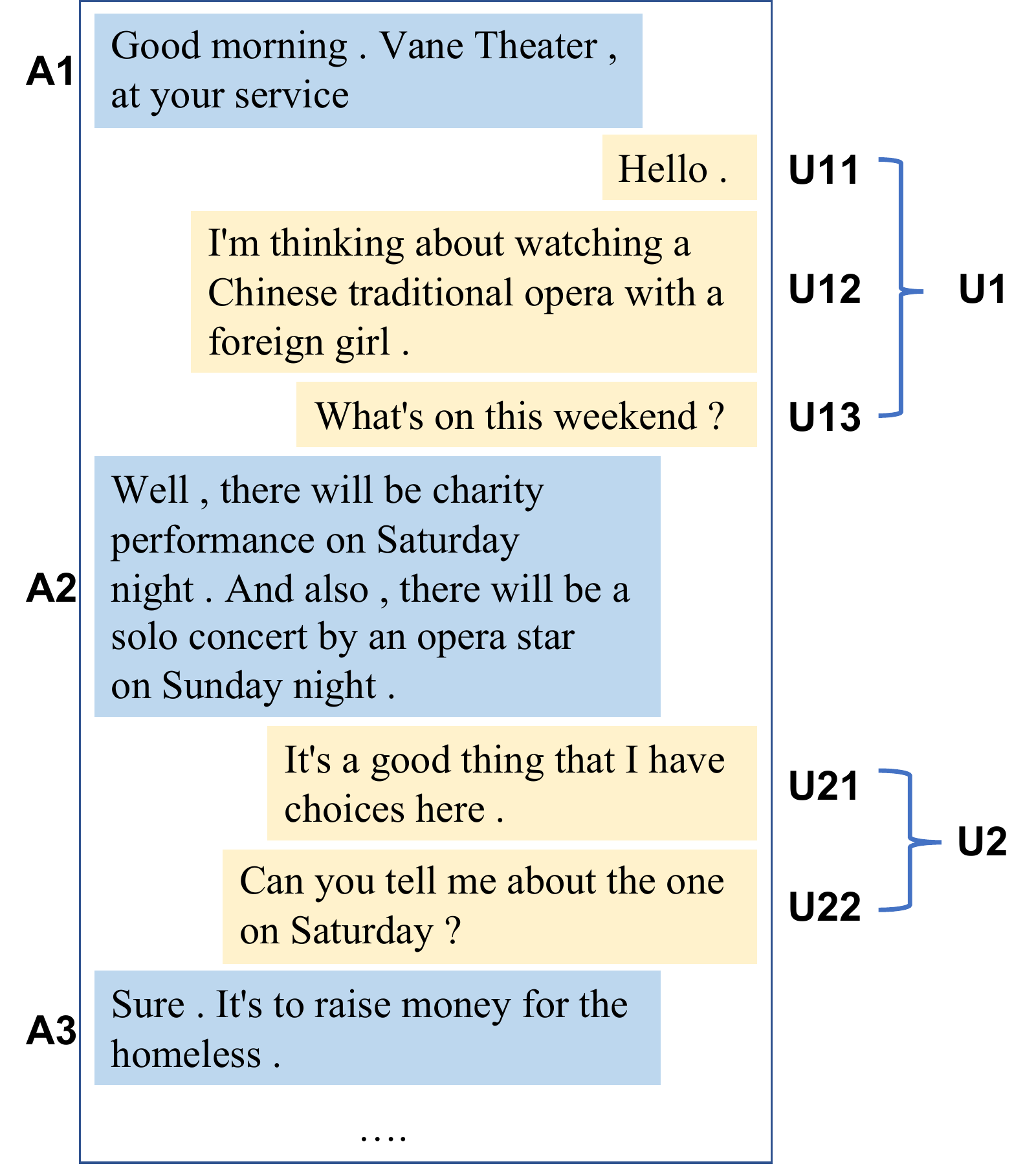}
\caption{A multi-turn dialogue fragment. In this case, user sends splited utterances in a turn, e.g. split U1 to \{U11, U12 and U13\}}
\label{fig:example_digloa}
\end{figure}

All species are unique, but languages make humans uniquest \cite{premack2004language}. Dialogues, especially spoken and written dialogues, are fundamental communication mechanisms for human beings. In real life, tons of businesses and entertainments are done via dialogues. This makes it significant and valuable to build an intelligent dialogue product. So far there are quite a few business applications of dialogue techniques, e.g. personal assistant, intelligent customer service and chitchat companion.

The quality of response is always the most important metric for dialogue agent, targeted by most existing work and models searching the best response. Some works incorporate knowledge \cite{DBLP:conf/acl/FungWM18,lin-etal-2019-task} to improve the success rate of task-oriented dialogue models, while some others \cite{NIPS2015_5866} solve the rare words problem and make response more fluent and informative.

Despite the heated competition of models, however, the pace of interaction is also important for human-computer dialogue agent, which has drawn less or no attention. Figure \ref{fig:example_digloa} shows a typical dialogue fragment in an instant message program. A user is asking the service about the schedule of the theater. The user firstly says hello~(\textbf{U11}) followed by demand description~(\textbf{U12}), and then asks for suggested arrangement~(\textbf{U13}), each of which is sent as a single message in one turn. The agent doesn't answer~(\textbf{A2}) until the user finishes his description and throws his question. The user then makes a decision~(\textbf{U21}) and asks a new question~(\textbf{U22}). And then the agent replies with~(\textbf{A3}). It's quite normal and natural that the user sends several messages in one turn and the agent waits until the user finished his last message, otherwise the pace of the conversation will be messed up. However, existing dialogue agents can not handle well when faced with this scenario and will reply to every utterance received immediately. 

There are two issues when applying existing dialogue agents to real life conversation. Firstly, when user sends a short utterance as the start of a conversation, the agent has to make a decision to avoid generating bad responses based on semantically incomplete utterance. Secondly, dialogue agent cutting in the conversation at an unreasonable time could confuse user and mess up the pace of conversation, leading to nonsense interactions. 

To address these two issues, in this paper, we propose a novel Imagine-then-Arbitrate (ITA) neural dialogue model to recognize if it is the appropriate moment for agent to reply when agent receives a message from the user. In our method, we have two imaginator modules and an arbitrator module. Imaginators will learn both of the agent's and user's speaking styles respectively. The arbitrator will use the dialogue history and the imagined future utterances generated by the two imaginators to decide whether the agent should wait user or make a response directly.

In summary, this paper makes the following contributions:
\begin{itemize}
    \item We first addressed an interaction problem, whether the dialogue model should wait for the end of the utterance or make a response directly in order to simulate real life conversation and tried several popular baseline models to solve it.
    \item We proposed a novel Imagine-then-Arbitrate (ITA) neural dialogue model to solve the problem mentioned above, based on both of the historical conversation information and the predicted future possible utterances.
    \item We modified two popular dialogue datasets to simulate the real human dialogue interaction behavior.
    \item Experimental results demonstrate that our model performs well on addressing ending prediction issue and the proposed imaginator modules can significantly help arbitrator outperform baseline models.
\end{itemize}

\section{Related Work}
\subsection{Dialogue System}
Creating a perfect artificial human-computer dialogue system is always a ultimate goal of natural language processing. In recent years, deep learning has become a basic technique in dialogue system. Lots of work has investigated on applying neural networks to dialogue system's components or end-to-end dialogue frameworks \cite{YanDCZZL17,lipton2018bbq-networks}. The advantage of deep learning is its ability to leverage large amount of data from internet, sensors, etc. The big conversation data and deep learning techniques like SEQ2SEQ \cite{NIPS2014_5346} and attention mechanism \cite{DBLP:conf/emnlp/LuongPM15} help the model understand the utterances, retrieve background knowledge and generate responses.

 \begin{figure*}
\centering
\includegraphics[width=1\linewidth]{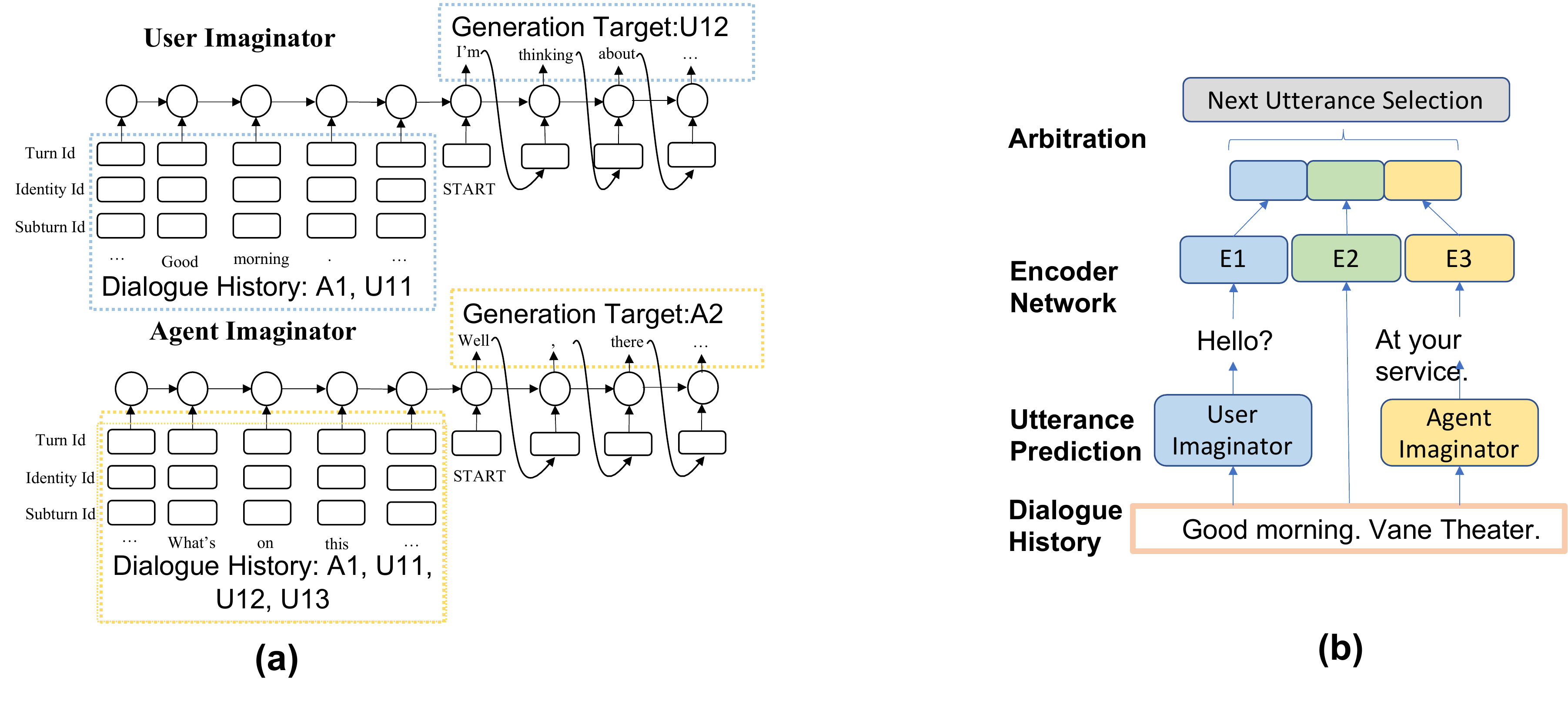}
\caption{Model Overview. (a) Train the agent and user imaginators using the same dialogues but different samples. (b) During training and inference step, arbitrator uses the dialogue history and two trained imaginators' predictions.}
\label{fig:model}
\end{figure*}

\subsection{Classification in Dialogue}

Though end-to-end methods play a more and more important role in dialogue system, the text classification modules \cite{jiang2018text,kowsari2017hdltex} remains very useful in many problems like emotion recognition \cite{song-etal-2019-generating}, gender recognition \cite{hoyle-etal-2019-unsupervised}, verbal intelligence, etc. There have been several widely used text classification methods proposed, e.g. Recurrent Neural Networks (RNNs) and CNNs. Typically RNN is trained to recognize patterns across time, while CNN learns to recognize patterns across space. \cite{kim2014convolutional} proposed TextCNNs trained on top of pre-trained word vectors for sentence-level classification tasks, and achieved excellent results on multiple benchmarks.

Besides RNNs and CNNs, \cite{vaswani2017attention} proposed a new network architecture called Transformer, based solely on attention mechanism and obtained promising performance on many NLP tasks. To make the best use of unlabeled data, \cite{devlin2018bert} introduced a new language representation model called BERT based on transformer and obtained state-of-the-art results.
\subsection{Dialogue Generation}
Different from retrieval method, Natural Language Generation (NLG) tries converting a communication goal, selected by the dialogue manager, into a natural language form. It reflects the naturalness of a dialogue system, and thus the user experience.
Traditional template or rule-based approach mainly contains a set of templates, rules, and hand-craft heuristics designed by domain experts. This makes it labor-intensive yet rigid, motivating researchers to find more data-driven approaches \cite{ghazvininejad2018knowledge, lin-etal-2019-task} that aim to optimize a generation module from corpora, one of which, Semantically Controlled LSTM (SC-LSTM) \cite{wen2015semantically}, a variant of LSTM \cite{hochreiter1997long}, gives a semantic control on language generation with an extra component. 

\section{Task Definition}
In this section we will describe the task by taking a scenario and then define the task formally.

As shown in Figure \ref{fig:example_digloa}, we have two participants in a conversation. One is the dialogue agent, and the other is a real human user. The agent's behavior is similar to most chatbots, except that it doesn't reply on every sentence received. Instead, this agent will judge to find the right time to reply.

Our problem is formulated as follows. There is a conversation history represented
as a sequence of utterances: $X = \{x_1, x_2, ..., x_m\}$, where
each utterance $x_i$ itself is a sequence
of words $x_{i_1}, x_{i_2}, x_{i_3}...x_{i_n}$. Besides, each utterance  has some additional tags:

\begin{itemize}
\item turn tags $t_0, t_1, t_2 ... t_k$ to show which turn this utterance is in the whole conversation. 
\item speakers' identification tags $agent$ or $user$ to show who sends this utterance. 
\item subturn tags ${st}_0, {st}_1, {st}_2 ... {st}_j$ for user to indicate which subturn an utterance $t_i$is in. Note that an utterance will be labelled as ${st}_0$ even if it doesn't have one.
\end{itemize}

Now, given a dialogue history $X$ and tags $T$, the goal of the model is to predict a label $Y \in \{0,1\}$, the action the agent would take, where $Y = 0$ means the agent will wait the user for next message, and $Y = 1$ means the agent will reply immediately. Formally we are going to maximize following probability:
\begin{equation}
    Y =\arg \max _{y} P\left(y | X, T\right)
\label{eq:problem}
\end{equation}


\section{Proposed Framework}
Basically, the task can be simplified as a simple text classification problem. However, traditional classification models only use the dialogue history $X$ and predict ground truth label. The ground truth label actually ignores all context information in the next utterance. To make the best use of training data, we propose a novel Imagine-then-Arbitrate (ITA) model taking $X$, ground truth label, and the future possible $X'$ into consideration. In this section, we will describe the architecture of our model and how it works in detail.


\subsection{Imaginator}
An imaginator is a natural language generator generating next sentence given the dialogue history. There are two imaginators in our method, agent's imaginator and user's imaginator. The goal of the two imaginators are to learn the agent’s and user’s speaking style respectively and generate possible future utterances.

As shown in Figure \ref{fig:model} \textbf{(a)}, imaginator itself is a sequence generation model. We use one-hot embedding to convert all words and relative tags, e.g. turn tags and place holders, to one-hot vectors $w_n \in \textbf{R}^V$, where $V$ is the length of vocabulary list. Then we extend each word $x_{i_j}$ in utterance $x_i$ by concatenating the token itself with turn tag, identity tag and subturn tag. We adopt SEQ2SEQ as the basic architecture and LSTMs as the encoder and decoder networks.  
LSTMs will encode each extended word $w_t$ as a continuous vector $h_t$ at each time step $t$. The process can be formulated as following:

\begin{equation}
\begin{array}{l}
{f_{t}=\sigma\left(W_{f} e(w_{t})+U_{f} h_{t-1}+b_{f}\right)} \\
{i_{t}=\sigma\left(W_{i} e(w_{t})+U_{i} h_{t-1}+b_{i}\right)} \\
{o_{t}=\sigma\left(W_{o} e(w_{t})+U_{o} h_{t-1}+b_{o}\right)} \\
{g_{t}=\tanh\left(W_{g} e(w_{t})+U_{g} h_{t-1}+b_{g}\right)} \\
c_{t}=f_{t} \odot c_{t-1}+i_{t} \odot g_{t} \\
h_{t}=o_{t} \odot \tanh \left(c_{t}\right)  \\
\end{array}\label{eq:lstm}
\end{equation}
where $e(w_t)$ is the embedding of the extended word $w_t$, $W_f$, $U_f$, $W_i$, $U_i$, $W_o$, $U_o$, $W_g$, $U_g$ and $b$ are learnt parameters. 

Though trained on the same dataset, the two imaginators learn different roles independently. So in the same piece of dialogue, we split it into different samples for different imaginators. For example, as shown in Figure \ref{fig:example_digloa} and \ref{fig:model} \textbf{(a)}, we use utterance (A1, U11, U12) as dialogue history input and U13 as ground truth to train the user imaginator and use utterance (A1, U11, U12, U13) as dialogue history and A2 as ground truth to train the agent imaginator.

During training, the encoder runs as equation \ref{eq:lstm}, and the decoder is the same structured LSTMs but $h_t$ will be fed to a Softmax with $W_{v} \in {\textbf{R}^{h \times V}}, b_{v} \in{\textbf{R}^\textbf{V}}$, which will produce a probability distribution $p_{t}$ over all words, formally:

\begin{equation}
\begin{array}{l}
p_{t}= Softmax(W_{v} h_{t}+b_{v})
\end{array}\label{eq:o_lstm}
\end{equation}

 the decoder at time step t will select the highest word in $p_{t}$, and our imaginator's loss is the sum of the negative log likelihood of the correct word at each step as follows:
\begin{equation}
L=-\sum_{t=1}^{N}\log(p_{t})
\end{equation}
where $N$ is the length of the generated sentence. During inference, we also apply beam search to improve the generation performance.

Finally, the trained agent imaginator and user imaginator are obtained.

\subsection{Arbitrator}
The arbitrator module is fundamentally a text classifier. However, in this task, we make the module maximally utilize both dialogue history and ground truth's semantic information. So we turned the problem of maximizing $Y$ from $X$ in equation (\ref{eq:problem}) to:
\begin{equation}
\begin{array}{l}
\begin{aligned}
    R_{agent} &= \textbf{IG}_{agent}(X, T) \\
    R_{user} &= \textbf{IG}_{user}(X, T) \\
    R' &=\arg \max _{y} P\left(y | X, T, R_{agent}, R_{user} \right) \\
    R' &\in {0, 1}
\end{aligned}
\end{array}\label{eq:new_problem}
\end{equation}
where $\textbf{IG}_{agent}$ and $\textbf{IG}_{user}$ are the trained agent imaginator and user imaginator respectively, and $R'$ is a selection indicator where $R' = 1$ means selecting $R_{agent}$ whereas $0$ means selecting $R_{user}$. And Thus we (1) introduce the generation ground truth semantic information and future possible predicted utterances (2) turn the label prediction problem into a response selection problem.

We adopt several architectures like Bi-GRUs, TextCNNs and BERT as the basis of arbitrator module. We will show how to build an arbitrator by taking TextCNNs as an example.

As is shown in Figure \ref{fig:model}, the three CNNs with same structure take the inferred responses $R_{agent}$, $R_{user}$ and dialogue history $X$, tags $T$. For each raw word sequence $x_1,...,x_n$, we embed each word as one-hot vector $w_{i} \in \textbf{R}^V$. By looking up a word embedding matrix $E \in \textbf{R}^{V \times d}$, the input text is represented as an input matrix $Q \in \textbf{R}^{l \times d}$, where $l$ is the length of sequence of words and $d$ is the dimension of word embedding features. The matrix is then fed into a convolution layer where a filter $\textbf{w} \in \textbf{R}^{k \times d}$ is applied: 
\begin{equation}
\begin{aligned}
c_{i} &= f (W \cdot Q_{i:i+k-1} + b)\\
\end{aligned}
\label{eq:conv}
\end{equation}
where $Q_{i:i+k-1}$ is the window of token representation and the function $f$ is $ReLU$, $W$ and $b$ are learnt parameters. Applying this filter to $m$ possible $Q_{i:i+k-1}$ obtains a feature map:

\begin{equation}
\begin{aligned}
\textbf{c} &= [c_1, c_2,.., c_{l-k+1}]
\end{aligned}
\label{eq:fmap}
\end{equation}
where $\textbf{c} \in \textbf{R}^{l-k+1}$
for $m$ filters. And we use $j \in \textbf{R} $ different size of filters in parallel in the same convolution layer. This means we will have $m_1, m_2, \dots, m_j$ windows at the same time, so formally:
\begin{equation}
\begin{aligned}
\textbf{C} &= [\textbf{c}_1, \textbf{c}_2,.., \textbf{c}_{j}]
\end{aligned}
\label{eq:ker}
\end{equation}

, then we apply max-over-time pooling operation to capture the most important feature:
\begin{equation}
\begin{aligned}
\hat{C} &= \max \{ \textbf{C} \}
\end{aligned}
\label{eq:max}
\end{equation}

, and thus we get the final feature map of the input sequence.

We apply same CNNs to get the feature maps of $X$, $R_{agent}$ and $R_{user}$:
\begin{equation}
\begin{aligned}
\hat{C}_{his}&=TextCNNs(X) \\
\hat{C}_{agent}&=TextCNNs(R_{agent}) \\
\hat{C}_{user}&=TextCNNs(R_{user}) \\
\end{aligned}
\end{equation}
where function TextCNNs() follows as equations from \ref{eq:conv} to \ref{eq:max}. Then we will have two possible dialogue paths, $X$ with $R_{agent}$ and $X$ with $R_{user}$, representations $D_{agent}$ and $D_{user}$:
\begin{equation}
\begin{aligned}
D_{agent} &= W_{1}[\hat{C}_{his}; \hat{C}_{agent}] + b_{1} \\
D_{user} &= W_{2}[\hat{C}_{his}; \hat{C}_{user}] + b_{2} \\
\end{aligned}
\end{equation}

And then, the arbitrator will calculate the probability of the two possible dialogue paths: 
\begin{equation}
\begin{aligned}
D &= W_{3}[D_{agent}; D_{user}] + b_{3} \\
P &= Softmax(W_{4}D+b_{4})
\end{aligned}
\end{equation}

Through learnt parameters $W_{4}$ and $b_{4}$, we will get a two-dimensional probability distribution $P$, in which the most reasonable response has the max probability. This also indicates whether the agent should wait or not. 

And the total loss function of the whole attribution module will be negative log likelihood of the probability of choosing the correct action:
\begin{equation}
    L = -\sum_{i=1}^{N}{\log(R_{i}'=Y_{i})}
\end{equation}
where $N$ is the number of samples and $Y_{i}$ is the ground truth label of i-th sample. 

The arbitrator module based on Bi-GRU and BERT is implemented similar to TextCNNs.
\begin{table*}[]
\centering
\begin{tabular}{|l|l|l|l|l|l|l|}
\hline
\multicolumn{1}{|c|}{Datasets}        & \multicolumn{3}{c|}{MultiWoz}                                                       & \multicolumn{3}{c|}{DailyDialogue}         \\ \hline
\multicolumn{1}{|c|}{}                & \multicolumn{1}{c|}{Train} & \multicolumn{1}{c|}{Valid} & \multicolumn{1}{c|}{Test} & \multicolumn{1}{c|}{Train} & Valid & Test  \\ \hline
\multicolumn{1}{|c|}{Vocabulary Size} & \multicolumn{3}{c|}{2443}                                                           & \multicolumn{3}{c|}{6219}                  \\ \hline
Dialogues                             & 8423                       & 1000                       & 1000                      & 1118                       & 1000  & 1000  \\ \hline
Avg. Turns/Dialogue                   & 6.32                       & 6.97                       & 6.98                      & 4.09                       & 4.21  & 4.03  \\ \hline
Avg. Split User Turns                 & 1.89                       & 1.92                       & 1.94                      & 2.09                       & 2.12  & 2.12  \\ \hline
Avg. Utterance Length                 & 10.54                      & 10.7                       & 10.56                     & 8.71                       & 8.54  & 8.75  \\ \hline
Avg. Agent's Utterance                & 14.43                      & 14.78                      & 14.69                     & 12.04                      & 11.81 & 12.17 \\ \hline
Avg. User's Utterance                 & 6.18                       & 6.28                       & 6.17                      & 5.91                       & 5.87  & 5.96  \\ \hline
Agent Wait Samples                    & 53249                      & 6970                       & 6983                      & 41547                      & 3846  & 3689  \\ \hline
Agent Reply Samples                   & 47341                      & 6410                       & 6573                      & 49540                      & 4717  & 4510  \\ \hline
\end{tabular}
\caption{Datasets Statistics. Note that the statistics are based on the modified dataset described in Section \ref{p:d_m}}
\label{tab:data_stastics}
\end{table*}

\begin{table*}[]
\centering
\begin{tabular}{|c|c|c|c|c|c|c|c|}
\hline
\multicolumn{2}{|c|}{Dataset}                                  & \multicolumn{3}{c|}{MultiWoz}                                                                                                           & \multicolumn{3}{c|}{DailyDialogue}                                                                                                      \\ \hline
\multicolumn{2}{|c|}{Task}                                     & Agent                                 & User                                  & Arbitrator                                             & Agent                               & User                                  & Arbitrator                                           \\ \hline
\multicolumn{2}{|c|}{Baseline: TextCNNs}                        & N/A                                   & N/A                                   & 77.68                                                   & N/A                                   & N/A                                   & 75.79                                                   \\ \hline
                                            & Agent Imaginator & \textbf{11.77}                        & 0.80                                  &                                                         & \textbf{4.51}                         & 0.61                                  &                                                         \\ \cline{2-4} \cline{6-7}
\multirow{-2}{*}{LSTM}                      & User Imaginator  & 0.3                                   & \textbf{8.87}                         & \multirow{-2}{*}{80.04}                                 & 0.15                                  & \textbf{8.70}                         & \multirow{-2}{*}{76.37}                                 \\ \hline
                                            & Agent Imaginator & \textbf{12.47}                        & 0.72                                  & {\color[HTML]{CB0000} }                                 & {\color[HTML]{CB0000} \textbf{19.19}} & 0.60                                  & {\color[HTML]{CB0000} }                                 \\ \cline{2-4} \cline{6-7}
\multirow{-2}{*}{LSTM+Attn.}                & User Imaginator  & 0.24                                  & \textbf{9.71}                         & \multirow{-2}{*}{{\color[HTML]{CB0000} \textbf{80.75}}} & 0.26                                  & \textbf{24.52}                        & \multirow{-2}{*}{{\color[HTML]{CB0000} \textbf{79.02}}} \\ \hline
                                            & Agent Imaginator & {\color[HTML]{CB0000} \textbf{13.37}} & 0.67                                  &                                                         & \textbf{19.01}                        & 0.67                                  &                                                         \\ \cline{2-4} \cline{6-7}
\multirow{-2}{*}{LSTM (with GLOVE) + Attn.} & User Imaginator  & 0.51                                  & {\color[HTML]{CB0000} \textbf{10.61}} & \multirow{-2}{*}{80.38}                                 & 0.21                                  & {\color[HTML]{CB0000} \textbf{24.65}} & \multirow{-2}{*}{78.56}                                 \\ \hline
\end{tabular}
\caption{Results of the different imaginators generation performance (in BLEU score) and accuracy score on the same TextCNNs based arbitrator. Better results between imaginators are in \textbf{BOLD} and best results on datasets are in \color[HTML]{CB0000} \textbf{RED}.}\label{tab:ab}
\end{table*}

\begin{table}[]\centering
\begin{tabular}{|l|l|l|}
\hline
Dataset                & MultiWoz                              & DailyDialogue                         \\ \hline
Random                 & 51.51                                 & 55.00                                 \\ \hline\hline
Bi-GRU                 & 79.12                                 & 75.23                                 \\ \hline
\textbf{ITA-GRU}     & \textbf{82.03}                        & \textbf{77.80}                        \\ \hline\hline
TextCNNs                & 77.68                                 & 75.79                                 \\ \hline
\textbf{ITA-TextCNN} & \textbf{80.75}                        & \textbf{79.02}                        \\ \hline \hline
BERT                   & 80.75                                 & 78.68                                 \\ \hline
\textbf{ITA-BERT}     & {\color[HTML]{CB0000} \textbf{82.73}} & {\color[HTML]{CB0000} \textbf{79.35}} \\ \hline
\end{tabular}
\caption{Accuracy Results on Two datasets. Better results between baselines and corresponding ITA models are in \textbf{BOLD} and best results on datasets are in \textcolor{red}{RED}. Random result is the accuracy of script that making random decisions.}
\label{tab:main_results}
\end{table}
\section{Experimental Setup}
\subsection{Datasets} \label{sec:datasets}
As the proposed approach mainly concentrates on the interaction of human-computer, we select and modify two very different style datasets to test the performance of our method. One is a task-oriented dialogue dataset MultiWoz 2.0 \deleted{\footnote{http://dialogue.mi.eng.cam.ac.uk/index.php/corpus/}}
 and the other  is a chitchat dataset DailyDialogue
 \deleted{\footnote{http://yanran.li/dailydialog.html}}. 
  Both datasets are collected from human-to-human conversations. We evaluate and compare the results with the baseline methods in multiple dimensions.  Table \ref{tab:data_stastics} shows the statistics of datasets.

\begin{itemize}
    \item \emph{MultiWOZ 2.0} \cite{budzianowski-etal-2018-multiwoz}. MultiDomain Wizard-of-Oz dataset (MultiWOZ) is a
    fully-labeled collection of human-human written conversations. Compared with previous task-oriented dialogue datasets, e.g. DSTC 2 \cite{henderson-etal-2014-second} and KVR \cite{DBLP:conf/sigdial/EricKCM17}, it is a much larger multi-turn conversational corpus\replaced{ and across serveral domains and topics}{: It is at least one order of magnitude larger than all previous annotated task-oriented corpora, with dialogues spanning across several domains and topics}.
    \item \emph{DailyDialogue} \cite{li-etal-2017-dailydialog}. DailyDialogue is a high-quality multi-turn dialogue dataset, which contains conversations about daily life. \deleted{In this dataset, humans often first respond to previous context and then propose their own questions and suggestions. In this way, people show their attention others’ words and are willing to continue the conversation.} Compare to the task-oriented dialogue datasets, the speaker's behavior will be more unpredictable and complex \deleted{for the arbitrator}.
\end{itemize}
\subsection{Datasets Modification}
\label{p:d_m}
Because the task we concentrate on is different from traditional ones, to make the datasets fit our problems and real life, we modify the datasets with the following steps:
\begin{table*}[]
\centering
\begin{tabular}{|l|l|}
\hline
\multicolumn{2}{|c|}{\textbf{Example}}                                                                                                                                                                                                                                                                                                                                                                       \\ \hline\hline
Dialogue History & \begin{tabular}[c]{@{}l@{}}User: actually \\ User: can you suggest {[}value\_count{]} of them\\ User: can i get their contact info as well\\ Agent: sure , i would suggest the {[}restaurant\_name{]} at {[}restaurant\_address{]} . you can reach them at \\\qquad\quad{[}restaurant\_phone{]} . i could reserve it for you\\ User: no\\ User: that ‘s ok\\ User: i can take it from here\end{tabular} \\ \hline
Ground Truth     & \textbf{User: thank,you for all your help }                                                                                                                                                                                                                                                                                                                                                        \\ \hline
Imagintator Prediction & \begin{tabular}[c]{@{}l@{}}Agent imaginator: would you like me to book it for you\\ User imaginator: thanks for all your help\end{tabular}    \\ \hline
Arbitrator Selection & \begin{tabular}[c]{@{}l@{}}\textbf{User imaginator}\end{tabular}                                                                                                                                                                                                                                                 \\ \hline

\end{tabular}\label{sample}
\caption{An Example of The Imaginator's Generation and arbitrator's Selection.}
\label{tab:example}
\end{table*}
\begin{itemize}
    \item \emph{Drop Slots and Values} For task-oriented dialogue, slot labels are important for navigating the system to complete a specific task. However, those labels and accurate values from ontology files will not benefit our task essentially. So we replace all specific values with a slot placeholder in preprocessing step.
    \item \emph{Split Utterances} Existing datasets concentrate on the dialogue content, combining multiple sentences into one utterance each turn when gathering the data. In this step, we randomly split the combined utterance into multiple utterances according to the punctuation. And we set a determined probability to decide if the preprocessing program should split a certain sentence.
    \item \emph{Add Turn Tag} We add turn tags, subturn tags and role tags to each split and original sentences to (1) label the speaker role and dialogue turns (2) tag the ground truth for training and testing the supervised baselines and our model.
\end{itemize}
Finally, we have the modified datasets which imitate the real life human chatting behaviors as shown in Figure \ref{fig:example_digloa}. Our datasets and code\deleted{\footnote{https://github.com/mumeblossom/ITA}} will be released to public for further researches in both academic and industry.

\subsection{Evaluation Method}
To compare with dataset baselines in multiple dimensions and test the model's performance, we use the overall Bilingual Evaluation Understudy (BLEU) \cite{DBLP:conf/acl/PapineniRWZ02} to evaluate the imaginators' generation performance. As for arbitrator, we use accuracy score of the classification to evaluate. Accuracy in our experiments is the correct ratio in all samples.

\subsection{Baselines and Training Setup}
The hyper-parameter settings adopted in baselines and our model are the best practice settings for each training set. All models are tested with various hyper-parameter settings to get their best performance. Baseline models are Bidirectional Gated Recurrent Units (Bi-GRUs) \cite{chung2014empirical}, TextCNNs \cite{kim2014convolutional} and BERT \cite{devlin2018bert}.



\section{Experimental Results and Analysis}
\subsection{Results}
In Table \ref{tab:ab}, we show different imaginators' generation abilities and their performances on the same TextCNN based arbitrator. Firstly, we gathered the results of agent and user imaginators' generation based on LSTM, LSTM-attention and LSTM-attention with GLOVE pretrained word embedding. According to the evaluation metric BLEU, the latter two models achieve higher but similar results. Secondly, when fixed the arbitrator on the TextCNNs model, the latter two also get the similar results on accuracy and significantly outperform the others including the TextCNNs baseline.

The performances on different arbitrators with the same LSTM-attention imaginators are shown in Table \ref{tab:main_results}. From those results, we can directly compared with the corresponding baseline models. The imaginators with BERT based arbitrator make the best results in both datasets while all ITA models beat the baseline models.

We also present an example of how our model runs in Table \ref{tab:example}. Imaginators predict the agent and user's utterance according to the dialogue history\deleted{(shown in model prediction)}, and then arbitrator selects the user imaginator's prediction\deleted{ that is more suitable with the dialogue history}. It is worth noting that the arbitrator generates a high-quality sentence again if only considering the generation effect. However, referring to the dialogue history,  it is not a good choice since its semantic is repeated in the last turn by the agent.

\subsection{Analysis}
\subsubsection{Imaginators Benefit the Performance} 
\deleted{From Table \ref{tab:main_results}, we can see that not only our BERT based model get the best results in both datasets, the other two models also significantly beat the corresponding baselines. Even the TextCNNs based model can beat all baselines in both datasets.  }

Table \ref{tab:ab} figures out experiment results on MultiWOZ dataset. The LSTM based agent imaginator get the BLEU score at 11.77 on agent samples, in which the ground truth is agents' utterances, and 0.80 on user samples. Meanwhile, the user imaginator get the BLEU score at 0.3 on agent samples and 8.87 on user target samples. Similar results are shown in other imaginators' expermients. Although these comparisons seem unfair to some extends since we do not have the agent and user's real utterances at the same time and under the same dialogue history, these results show that the imaginators did learn the speaking style of agent and user respectively. So the suitable imaginator's generation will be more similar to the ground truth, such an example shown in Table \ref{tab:example}, which means this response more semantically suitable given the dialogue history. 

If we fix the agent and user imaginators' model, as we take the LSTM-attention model, the arbitrators achieve different performances on different models, shown in Table~\ref{tab:main_results}. As expected, ITA models beat their base models by nearly 2 $\sim$ 3\% and ITA-BERT model beats all other ITA models.

So from the all results, we can conclude that imaginators will significantly help the arbitrator in predicting the dialogue interaction behavior using the future possible agent and user responses’ semantic information.

\subsubsection{Relation of Imaginators and Arbitrator's Performance} 
As shown in the DailyDialogue dataset of Table \ref{tab:ab}, we can see that attention mechanism works in learning the generation task. LSTMs -Attention and LSTMs-attention-GLOVE based imaginators get more than 19 and 24 BLEU scores in corresponding target, while the LSTMs without attention gets only 4.51 and 8.70. These results also impact on the arbitrator results. The imaginator with attention mechanism get an accuracy score of 79.02 and 78.56, significantly better than the others. The evidence also exists in the results on MultiWoz. All imaginators get similar generation performance, so the arbitrators gets the similar accuracy scores.  

From those results, we can conclude that there is positive correlation between the performance of imaginators and arbitrators. However, there still exists problems. It's not easy to evaluate the dialogue generation's performance. In the results of MultiWoz, we can see that LSTMs-GLOVE based ITA performs a little better than LSTMs-attention based ITA, but not the results of the arbitrator are opposite. This may indicate that (1) when the imaginators' performance is high enough, the arbitrator's performance will be stable and (2) the BLEU score will not perfectly present the contribution to the arbitrator. We leave these hypotheses in future work.




\section{Conclusion}
We first address an interaction problem, whether the dialogue model should wait for the end of the utterance or reply directly in order to simulate user's real life conversation behavior, and propose a novel Imagine-then-Arbitrate (ITA) neural dialogue model to deal with it. Our model introduces the imagined future possible semantic information for prediction. We modified two popular dialogue datasets to fit in the real situation. It is reasonable that additional information is helpful for arbitrator, despite its fantasy.

\bibliographystyle{named}
\bibliography{ijcai20}

\begin{thebibliography}{}

\bibitem[\protect\citeauthoryear{Budzianowski \bgroup \em et al.\egroup
  }{2018}]{budzianowski-etal-2018-multiwoz}
Pawe{\l} Budzianowski, Tsung-Hsien Wen, Bo-Hsiang Tseng, I{\~n}igo Casanueva,
  Stefan Ultes, Osman Ramadan, and Milica Ga{\v{s}}i{\'c}.
\newblock {M}ulti{WOZ} - a large-scale multi-domain wizard-of-{O}z dataset for
  task-oriented dialogue modelling.
\newblock In {\em Proceedings of the 2018 EMNLP}, pages 5016--5026, Brussels,
  Belgium, October-November 2018. ACL.

\bibitem[\protect\citeauthoryear{Chung \bgroup \em et al.\egroup
  }{2014}]{chung2014empirical}
Junyoung Chung, Caglar Gulcehre, Kyunghyun Cho, and Yoshua Bengio.
\newblock Empirical evaluation of gated recurrent neural networks on sequence
  modeling.
\newblock In {\em NIPS 2014 Workshop on Deep Learning}, 2014.

\bibitem[\protect\citeauthoryear{Devlin \bgroup \em et al.\egroup
  }{2018}]{devlin2018bert}
Jacob Devlin, Ming-Wei Chang, Kenton Lee, and Kristina Toutanova.
\newblock Bert: Pre-training of deep bidirectional transformers for language
  understanding.
\newblock {\em arXiv preprint arXiv:1810.04805}, 2018.

\bibitem[\protect\citeauthoryear{Eric and
  Manning}{2017}]{DBLP:conf/sigdial/EricKCM17}
Mihail Eric and Christopher~D Manning.
\newblock Key-value retrieval networks for task-oriented dialogue.
\newblock {\em arXiv preprint arXiv:1705.05414}, 2017.

\bibitem[\protect\citeauthoryear{Ghazvininejad \bgroup \em et al.\egroup
  }{2018}]{ghazvininejad2018knowledge}
Marjan Ghazvininejad, Chris Brockett, Ming-Wei Chang, Bill Dolan, Jianfeng Gao,
  Wen-tau Yih, and Michel Galley.
\newblock A knowledge-grounded neural conversation model.
\newblock In {\em Thirty-Second AAAI}, 2018.

\bibitem[\protect\citeauthoryear{Henderson \bgroup \em et al.\egroup
  }{2014}]{henderson-etal-2014-second}
Matthew Henderson, Blaise Thomson, and Jason~D. Williams.
\newblock The second dialog state tracking challenge.
\newblock In {\em Proceedings of the 15th SIGDIAL}, pages 263--272,
  Philadelphia, PA, U.S.A., June 2014. ACL.

\bibitem[\protect\citeauthoryear{Hochreiter and
  Schmidhuber}{1997}]{hochreiter1997long}
Sepp Hochreiter and J{\"u}rgen Schmidhuber.
\newblock Long short-term memory.
\newblock {\em Neural computation}, 9(8):1735--1780, 1997.

\bibitem[\protect\citeauthoryear{Hoyle \bgroup \em et al.\egroup
  }{2019}]{hoyle-etal-2019-unsupervised}
Alexander~Miserlis Hoyle, Lawrence Wolf-Sonkin, Hanna Wallach, Isabelle
  Augenstein, and Ryan Cotterell.
\newblock Unsupervised discovery of gendered language through latent-variable
  modeling.
\newblock In {\em Proceedings of the 57th ACL}, pages 1706--1716, Florence,
  Italy, July 2019. ACL.

\bibitem[\protect\citeauthoryear{Jiang \bgroup \em et al.\egroup
  }{2018}]{jiang2018text}
Mingyang Jiang, Yanchun Liang, Xiaoyue Feng, Xiaojing Fan, Zhili Pei, Yu~Xue,
  and Renchu Guan.
\newblock Text classification based on deep belief network and softmax
  regression.
\newblock {\em Neural Computing and Applications}, 29(1):61--70, 2018.

\bibitem[\protect\citeauthoryear{Kim}{2014}]{kim2014convolutional}
Yoon Kim.
\newblock Convolutional neural networks for sentence classification.
\newblock In {\em Proceedings of the 2014 EMNLP}, pages 1746--1751, 2014.

\bibitem[\protect\citeauthoryear{Kowsari \bgroup \em et al.\egroup
  }{2017}]{kowsari2017hdltex}
Kamran Kowsari, Donald~E Brown, Mojtaba Heidarysafa, Kiana~Jafari Meimandi,
  Matthew~S Gerber, and Laura~E Barnes.
\newblock Hdltex: Hierarchical deep learning for text classification.
\newblock In {\em 2017 ICMLA}, pages 364--371. IEEE, 2017.

\bibitem[\protect\citeauthoryear{Li \bgroup \em et al.\egroup
  }{2017}]{li-etal-2017-dailydialog}
Yanran Li, Hui Su, Xiaoyu Shen, Wenjie Li, Ziqiang Cao, and Shuzi Niu.
\newblock {D}aily{D}ialog: A manually labelled multi-turn dialogue dataset.
\newblock In {\em Proceedings of the 8th IJCNLP}, pages 986--995, Taipei,
  Taiwan, November 2017. Asian Federation of Natural Language Processing.

\bibitem[\protect\citeauthoryear{Lin \bgroup \em et al.\egroup
  }{2019}]{lin-etal-2019-task}
Zehao Lin, Xinjing Huang, Feng Ji, Haiqing Chen, and Yin Zhang.
\newblock Task-oriented conversation generation using heterogeneous memory
  networks.
\newblock In {\em Proceedings of the 2019 EMNLP-IJCNLP}, pages 4557--4566, Hong
  Kong, China, November 2019. ACL.

\bibitem[\protect\citeauthoryear{Lipton \bgroup \em et al.\egroup
  }{2018}]{lipton2018bbq-networks}
Zachary Lipton, Xiujun Li, Jianfeng Gao, Lihong Li, Faisal Ahmed, and Li~Deng.
\newblock Bbq-networks: Efficient exploration in deep reinforcement learning
  for task-oriented dialogue systems.
\newblock In {\em AAAI 2018}, February 2018.

\bibitem[\protect\citeauthoryear{Luong \bgroup \em et al.\egroup
  }{2015}]{DBLP:conf/emnlp/LuongPM15}
Thang Luong, Hieu Pham, and Christopher~D. Manning.
\newblock Effective approaches to attention-based neural machine translation.
\newblock In {\em Proceedings of the {EMNLP} 2015, Lisbon, Portugal}, pages
  1412--1421, 2015.

\bibitem[\protect\citeauthoryear{Madotto \bgroup \em et al.\egroup
  }{2018}]{DBLP:conf/acl/FungWM18}
Andrea Madotto, Chien-Sheng Wu, and Pascale Fung.
\newblock Mem2seq: Effectively incorporating knowledge bases into end-to-end
  task-oriented dialog systems.
\newblock In {\em Proceedings of the 56th ACL}, pages 1468--1478, 2018.

\bibitem[\protect\citeauthoryear{Papineni \bgroup \em et al.\egroup
  }{2002}]{DBLP:conf/acl/PapineniRWZ02}
Kishore Papineni, Salim Roukos, Todd Ward, and Wei{-}Jing Zhu.
\newblock Bleu: a method for automatic evaluation of machine translation.
\newblock In {\em Proceedings of the 40th ACL, July 6-12, 2002, Philadelphia,
  PA, {USA.}}, pages 311--318, 2002.

\bibitem[\protect\citeauthoryear{Premack}{2004}]{premack2004language}
David Premack.
\newblock Is language the key to human intelligence?
\newblock {\em Science}, 303(5656):318--320, 2004.

\bibitem[\protect\citeauthoryear{Song \bgroup \em et al.\egroup
  }{2019}]{song-etal-2019-generating}
Zhenqiao Song, Xiaoqing Zheng, Lu~Liu, Mu~Xu, and Xuanjing Huang.
\newblock Generating responses with a specific emotion in dialog.
\newblock In {\em Proceedings of the 57th ACL}, pages 3685--3695, Florence,
  Italy, July 2019. ACL.

\bibitem[\protect\citeauthoryear{Sutskever \bgroup \em et al.\egroup
  }{2014}]{NIPS2014_5346}
Ilya Sutskever, Oriol Vinyals, and Quoc~V. Le.
\newblock Sequence to sequence learning with neural networks.
\newblock In {\em NIPS 2014, December 8-13 2014, Montreal, Quebec, Canada},
  pages 3104--3112, 2014.

\bibitem[\protect\citeauthoryear{Vaswani \bgroup \em et al.\egroup
  }{2017}]{vaswani2017attention}
Ashish Vaswani, Noam Shazeer, Niki Parmar, Jakob Uszkoreit, Llion Jones,
  Aidan~N Gomez, {\L}ukasz Kaiser, and Illia Polosukhin.
\newblock Attention is all you need.
\newblock In {\em Advances in neural information processing systems}, pages
  5998--6008, 2017.

\bibitem[\protect\citeauthoryear{Vinyals \bgroup \em et al.\egroup
  }{2015}]{NIPS2015_5866}
Oriol Vinyals, Meire Fortunato, and Navdeep Jaitly.
\newblock Pointer networks.
\newblock In {\em Advances in Neural Information Processing Systems 28: Annual
  Conference on Neural Information Processing Systems 2015, December 7-12,
  2015, Montreal, Quebec, Canada}, pages 2692--2700, 2015.

\bibitem[\protect\citeauthoryear{Wen \bgroup \em et al.\egroup
  }{2015}]{wen2015semantically}
Tsung-Hsien Wen, Milica Gasic, Nikola Mrksic, Pei-Hao Su, David Vandyke, and
  Steve Young.
\newblock Semantically conditioned lstm-based natural language generation for
  spoken dialogue systems.
\newblock {\em arXiv preprint arXiv:1508.01745}, 2015.

\bibitem[\protect\citeauthoryear{Yan \bgroup \em et al.\egroup
  }{2017}]{YanDCZZL17}
Zhao Yan, Nan Duan, Peng Chen, Ming Zhou, Jianshe Zhou, and Zhoujun Li.
\newblock Building task-oriented dialogue systems for online shopping.
\newblock In {\em Proceedings of the Thirty-First {AAAI} Conference on
  Artificial Intelligence, February 4-9, 2017, San Francisco, California,
  {USA.}}, pages 4618--4626, 2017.

\end{thebibliography}

\end{document}